\pdfoutput=1

\documentclass[11pt]{article}

\usepackage[]{acl}

\usepackage{times}
\usepackage{latexsym}

\usepackage{amsmath}
\usepackage{amsfonts}
\usepackage{booktabs}

\def\eflomal{Eflomal}

\usepackage[T1]{fontenc}

\usepackage[utf8]{inputenc}

\usepackage[inline]{enumitem}
\usepackage{microtype}
\usepackage{graphicx}
\usepackage{subfig}
\usepackage{caption}

\usepackage{etoolbox}

 \def\tablabel#1{\label{tab:#1}\label{p:#1}}

\def\myblue#1{#1}

\newcounter{notecounter}
\newcommand{\enotesoff}{\long\gdef\enote##1##2{}}
\newcommand{\enoteson}{\long\gdef\enote##1##2{{
\stepcounter{notecounter}=
{\large\bf
\hspace{1cm}\arabic{notecounter} $<<<$ ##1: ##2
$>>>$\hspace{1cm}}}}}

\enoteson
\enotesoff

\title{Graph Neural Networks for Multiparallel Word Alignment}

\author{Ayyoob Imani$^1$,
L\"{u}tfi Kerem \c{S}enel$^1$, Masoud Jalili Sabet$^1$, \\
	\textbf{{Fran\c{c}ois Yvon$^2$, Hinrich Sch\"{u}tze$^1$}}\\
	$^1$Center for Information and Language Processing (CIS), LMU Munich, Germany\\
	$^2$Universit\'{e} Paris-Saclay, CNRS, LISN, France\\
	{\tt \{ayyoob, masoud, lksenel\}@cis.lmu.de,} \\
	{\tt francois.yvon@limsi.fr}
}

\def\tablabel#1{\label{tab:#1}\label{p:#1}}

\def\secref#1{\S\ref{sec:#1}}
\def\seclabel#1{\label{sec:#1}}

\begin{document}
\maketitle
\begin{abstract}
  After a period of decrease, interest in word alignments is
  increasing again for their usefulness
  in domains  such as typological research, 
  cross-lingual annotation projection and 
  machine translation. Generally, alignment algorithms 
  only use  bitext and do not make use of the fact that 
  many parallel corpora are multiparallel. 
  \myblue{Here, we compute high-quality word alignments 
  between multiple language pairs by considering all language pairs together.}
  \myblue{First, we create a multiparallel word alignment graph,
   joining all bilingual word alignment pairs in one graph.
  Next, we use 
  graph neural networks (GNNs) to exploit the graph structure.}
  Our GNN approach (i) utilizes information about the meaning, 
  position and language of the input words, (ii) incorporates 
  information from multiple parallel sentences,
  (iii) \myblue{ adds and removes edges from the initial alignments,} and
  (iv) yields a prediction model that can generalize beyond 
  the training sentences.  We show that community detection 
  provides valuable information for multiparallel word alignment.
  Our method outperforms previous work on three word alignment datasets
  and on a downstream task.
\end{abstract}

\section{Introduction}
Word alignments are crucial for statistical machine translation 
\citep{koehn2003statistical}
and useful for many other multilingual tasks such as
neural machine translation 
\citep{alkhouli-ney-2017-biasing, alkhouli-etal-2016-alignment},
typological analysis \citep{lewis-xia-2008-automatically,
ostling-2015-word, asgari-schutze-2017-past} and
annotation projection \citep{yarowsky-ngai-2001-inducing, fossum-abney-2005-automatically, 
wisniewski-etal-2014-cross, huck-etal-2019-cross}.
The rise of deep learning initially led to a temporary plateau, 
but interest in word alignments is now increasing, demonstrated by 
several recent publications \citep{jalili-sabet-etal-2020-simalign,chen-etal-2020-accurate,dou-neubig-2021-word}.

\myblue{While word alignment is usually considered for bilingual corpora, our work addresses
  the problem of \emph{word alignment in multiparallel corpora}, containing sentence
  level parallel text in more than two languages.
Examples of multiparallel corpora are JW300 \cite{agic2019jw300}, 
PBC \cite{mayer2014creating} which covers the highest number of languages  ($1334$),
and Tatoeba.\footnote{\url{https://tatoeba.org}}
While the per-language amount of data provided in such corpora is less than
bilingual corpora, they support highly low-resource languages, many of which are not covered by existing
language technologies \cite{joshi2020state}.} 
Therefore, these corpora are essential for studying many of the world's 
low-resource languages.

\begin{figure}
  \centering
  \includegraphics[width=0.9\columnwidth]{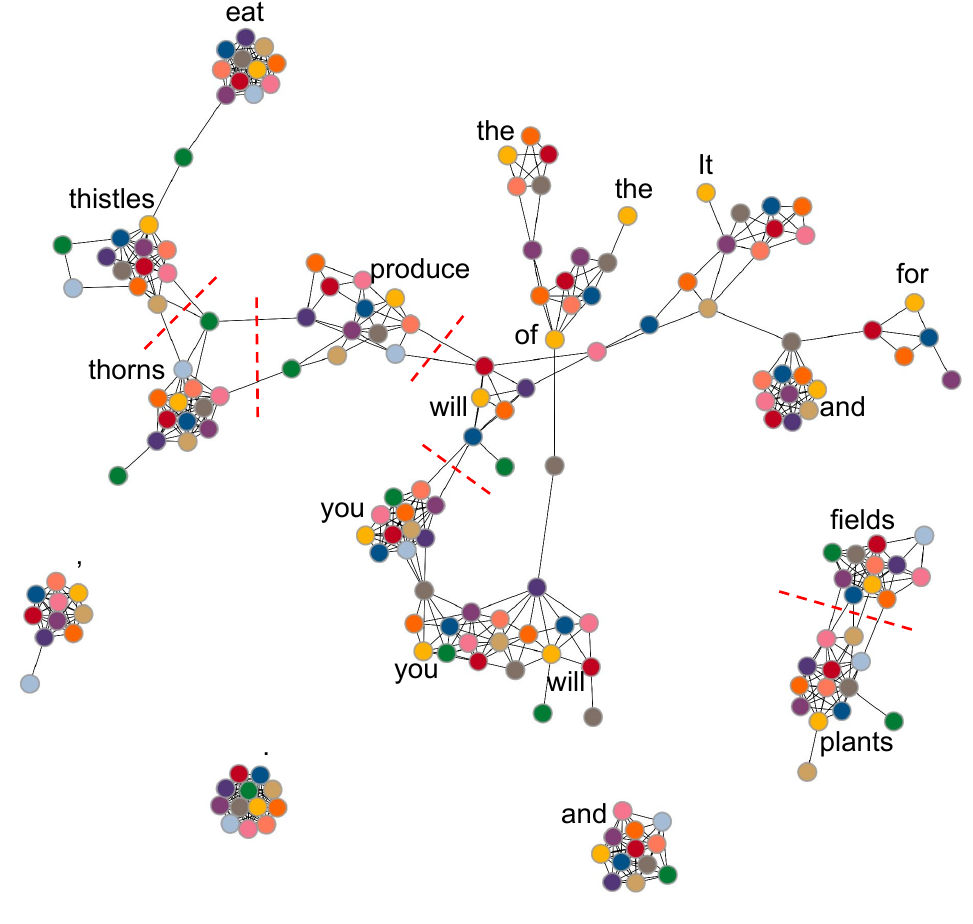}
  \caption{Alignment graph for the verse 
  ``It will produce thorns and thistles for you, 
  and you will eat the plants of the field.''\ in a 12-way multiparallel corpus. 
  Colors represent  languages. Each English (yellow) node is
  annotated with its word.
  Red dashed lines cut links that incorrectly connect 
  distinct concepts. We exploit 
  community detection algorithms \myblue{to detect distinct
  concepts. This provides valuable information for our GNN model
  and improves word alignments.}}
   \label{fig:alignment_graph}
\end{figure}

We consider the task of word alignment for 
multiparallel sentences.  The basic motivation is that the
alignment between words in languages $U$ and $V$ can benefit
from word-level alignments of $U$ and $V$ with a translation
in a third language $W$.  Following up on the work
of \citet{imani2021MPWA}, we model multilingual word
alignments with tools borrowed from graph theory (community
detection algorithms) combined with neural network based
models, specifically, the graph neural network (GNN) model
of~\citet{Scarselli2009GNN}.

GNNs were proposed to extend the powerful current generation of 
neural network models to the processing of graph-structured data and
they have gained increasing popularity in many domains
 \citep{wu2020graphSocial, sanchez18GNNPhysics, he2020GNNrecommender}. 
 GNNs can incorporate heterogeneous sources
of signal in the form of node and edge features. We use this property  to
take into account properties of the whole alignment graph, notably its tendency to cluster
into \emph{communities}, see Figure~\ref{fig:alignment_graph}.

With our new proposed methods, we obtain improved results
on word alignment for three language pairs:
English-French,
Finnish-Hebrew and Finnish-Greek.
As a demonstration of the importance of high-quality alignments,
we use our word alignments to project annotations from high-resource  to low-resource languages.
We improve a part-of-speech tagger for Yoruba by
training it over a high-quality dataset, which is created using annotation projection.

\textbf{Contributions:}
\begin{enumerate*}
  [label={\textbf{\roman{*})}}] 
  \item \myblue{We propose a graph neural network model to enhance word alignments 
  in a multiparallel corpus. The model 
  incorporates a diverse set of features 
for word alignments in multiparallel corpora and 
an elegant way of training it efficiently and effectively.}
  \item We show that community detection  improves
multiparallel word alignment. 
  \item We show that the improved alignments improve performance
  on a downstream task \myblue{for a low resource language.}
  \item We propose a new method to infer alignments from the alignment probability matrix.
  \item We will make our code publicly available.
\end{enumerate*}

\section{MultiParallel Word Alignment Graphs}

\subsection{Building MultiParallel Word Alignment Graphs}

Our starting point is the work of \citet{imani2021MPWA}, who introduce 
MPWA (MultiParallel Word Alignment), a framework that utilizes the synergy 
between multiple language pairs to improve bilingual word alignments for a target language pair. 
The rationale is that some of the missing alignment edges between a source and a target language 
can be recovered using their alignments with words in other languages. 

An MPWA graph is constructed using the following two steps:
\begin{enumerate}
  \item create initial bilingual alignments
  for all language pairs in a multiparallel corpus using a bilingual word aligner;
  \item represent the bilingual alignments for each multiparallel sentence in a graph containing one vertex for each token occurring in any language and one edge for each initial bilingual word alignment link.
\end{enumerate}
Potentially missing alignment links are then added
based on the graph structure in an inference step, casting
the word alignment task as an edge prediction problem.
Figure~\ref{fig:alignment_graph} gives an example of a multiparallel word alignment graph for a 12-way multiparallel sentence.

\citet{imani2021MPWA} use two traditional graph algorithms, Adamic-Adar and 
non-negative matrix factorization, for predicting new alignment edges from the MPWA graph.
However, these graph algorithms are applied to individual multiparallel
sentences independently and therefore cannot accumulate knowledge from
multiple sentences. Moreover, their edge predictions are solely 
based on the structure of the graph and do not take advantage of other 
beneficial signals such as a word's language, 
relative position and meaning.
Another limitation of this work is that it 
only adds links and does not remove any, which is important to improve precision.

This work addresses these shortcomings by using GNNs to predict alignment edges from MPWA graphs.

\subsection{\myblue{Community Detection in Alignment Graphs}}
\seclabel{ssec:community}
One important advantage of GNNs over traditional graph algorithms is that they can directly incorporate signals from different sources in the form of node and edge features. We utilize this by taking into account the following observation:
The nodes in the alignment graph
are words in parallel sentences that are translations of each other.
If the initial bilingual alignments are of good quality, we expect
words that are mutual translations to form densely connected regions
or \emph{communities};
see Figure~\ref{fig:alignment_graph}. These communities
should not be linked to each other,
each corresponding to a distinct connected component. In other words, ideally, 
words representing a concept should be densely connected,
but there should be no links between different concepts. 
\myblue{While, this intuition will not be true for all concepts between all possible language pairs, we nonetheless hypothesize that identifying distinct concepts in a multiparallel word alignment graph can provide useful information.}

To examine to what extent these expectations are met,
we count the components in the original
Eflomal-generated
\cite{ostling2016efficient}
graph (see \secref{ssec:init_aligns} for details on the initial alignments).
Table~\ref{tab:communities} shows that 
the average number of components per sentence is less than
three (``Eflomal intersection'', columns \#CC).
But intuitively, the number of components
should roughly correspond to sentence length (or, more
precisely, 
the number of content words). This indicates that there
are many links that incorrectly connect different concepts.
To detect such links, we use community detection (CD) algorithms.

CD algorithms find subnetworks of nodes that form tightly knit
groups that are only loosely connected with a small number
of links \cite{girvan2002community}. One well-known approach
to CD attempts to maximize
the modularity measure \cite{newman2004finding}. Modularity
assesses how beneficial a division of a community into two
communities is, in the sense that there are many links within
communities and only a few between them.
Given a graph $G$ with $n$ nodes and $m$ edges
and $G$'s adjacency matrix $A\in \mathbb{R}^{n \times n} $,
modularity is defined as:
\begin{equation}
  mod = \frac{1}{2m} \sum_{ij} \left( A_{ij} - \gamma\frac{d_id_j}{2m}\right)
  I(c_i,c_j)
\end{equation}
where $d_i$ is the degree of node $i$.  $I(c_i,c_j)$ is $1$ if 
nodes $i$ and
$j$ are in the same community, $0$ otherwise.

As exact modularity maximization is intractable, we experiment with
two CD algorithms implementing different heuristic approaches:
\begin{itemize}
  \item Greedy modularity communities (GMC).
  This method uses Clauset-Newman-Moore greedy modularity
  maximization \cite{clauset2004finding}. GMC
  begins with each node in its own community and greedily joins the pair
  of communities that most increases modularity until no such pair exists.
  
  \item Label propagation communities (LPC).
  This method finds communities in a graph using
  label propagation  \citep{cordasco2010community}. 
  It begins by giving a label to each node of the network.
  Then each node's label is updated by the most frequent
  label among its neighbors in each iteration.  
  It performs label propagation on a portion of nodes at each step and
  quickly converges to a stable labeling.
\end{itemize}

After detecting communities, we link all nodes inside a community and
remove all inter-community links. GMC (LPC) on average
removes 3\% (7\%) of the edges.
Table~\ref{tab:communities} reports
the average number of graph components per sentence
before and after running GMC and LPC, as well as the corresponding $F_1$ for word alignment (see \secref{ssec:align_datasets} for details on the evaluation datasets).
We see that  the number of communities found 
is lower for GMC than for LPC; therefore, 
LPC identifies more candidate links for
deletion.\footnote{LPC may detect more communities than average
sentence length because of null words:
words
that have no translation in the other languages, giving rise
to separate communities.
}
Comparing the number
of communities detected with the average sentence length, 
GMC 
seems to have failed to detect enough communities to split different concepts properly.
The $F_1$ scores confirm this observation and show that
LPC performs well at detecting the communities we
are looking for.

This analysis shows that CD algorithms
compute valuable information for word alignments.
To exploit this in our GNN model, we add node
community information
as a node feature; see \secref{sssec:nodefeatures}.

\begin{table}
  \scriptsize
	\centering
	\def\tablesep{0.2cm}
  \begin{tabular}{
  @{\hspace{\tablesep}}l@{\hspace{\tablesep}}||
  @{\hspace{\tablesep}}r@{\hspace{\tablesep}}r@{\hspace{\tablesep}}
  @{\hspace{\tablesep}}r@{\hspace{\tablesep}}r@{\hspace{\tablesep}}
  @{\hspace{\tablesep}}r@{\hspace{\tablesep}}r@{\hspace{\tablesep}}}
 
   & \multicolumn{2}{c}{FIN-HEB} & \multicolumn{2}{c}{FIN-GRC} & \multicolumn{2}{c}{ENG-FRA} \\ 
& \#CC & $F_1$
& \#CC & $F_1$
& \#CC & $F_1$\\
\midrule \midrule
  Eflomal intersection & 2.2 & 0.404 & 1.6 & 0.646 & 2.2 & 0.678 \\
  \midrule
  GMC & 13.7 & 0.396 & 10.1 & 0.375 & 13.5 & 0.411 \\ 
  LPC & 41.5 & 0.713 & 37.1 & 0.754 & 46.0 & 0.767 \\ 
  \midrule
  Sentence length & \multicolumn{2}{c}{25.7} & \multicolumn{2}{c}{23.2} & \multicolumn{2}{c}{27.4} \\
  \midrule \midrule
  \end{tabular}
  \caption{Effect of community detection algorithms
(GMC and LPC)
on
  alignment prediction.
\#CC: average number of connected components. $F_1$:
word alignment performance.
  }
  \label{tab:communities}
\end{table}

\section{Predicting and using
MultiParallel Word Alignments (MPWAs)}
\subsection{GNNs for MPWA}
\seclabel{ssec:gnns_in_mpwa}

GNNs can be used in transductive or inductive settings. Transductively,
the final model can only be used
for inference over the same graph that it is trained on. 
In an inductive setting, which we use here,
nodes are represented as feature vectors, and the final model
has the advantage of being applicable to a different graph in inference.

Below is the step-by-step overview of our GNN-based approach for an MPWA graph:
\begin{enumerate}
\item run community detection algorithms on the initial graph (\secref{ssec:community});
\item obtain features for the nodes of the graph (\secref{sssec:nodefeatures});
\item compute node embeddings from node features and initial alignment links in the GNN encoding step (\secref{sssec:gnn_training});
\item learn to distinguish between nodes that are aligned together and that are not aligned together in the GNN decoding step (\secref{sssec:gnn_training});
\end{enumerate}
After the GNN model is trained on multiple MPWA graphs,
it is used to infer an alignment probability matrix between tokens in a source language and tokens in a target language for a multiparallel sentence, see \secref{sssec:inducing_edges}.
Our method predicts new alignment links from this matrix,
independently of initial edges. Therefore,
given an initial bilingual alignment,
it is not limited to adding edges, but it can also remove them.

\subsubsection{Model Architecture}

\begin{figure*}
  \centering
  \includegraphics[width=\textwidth]{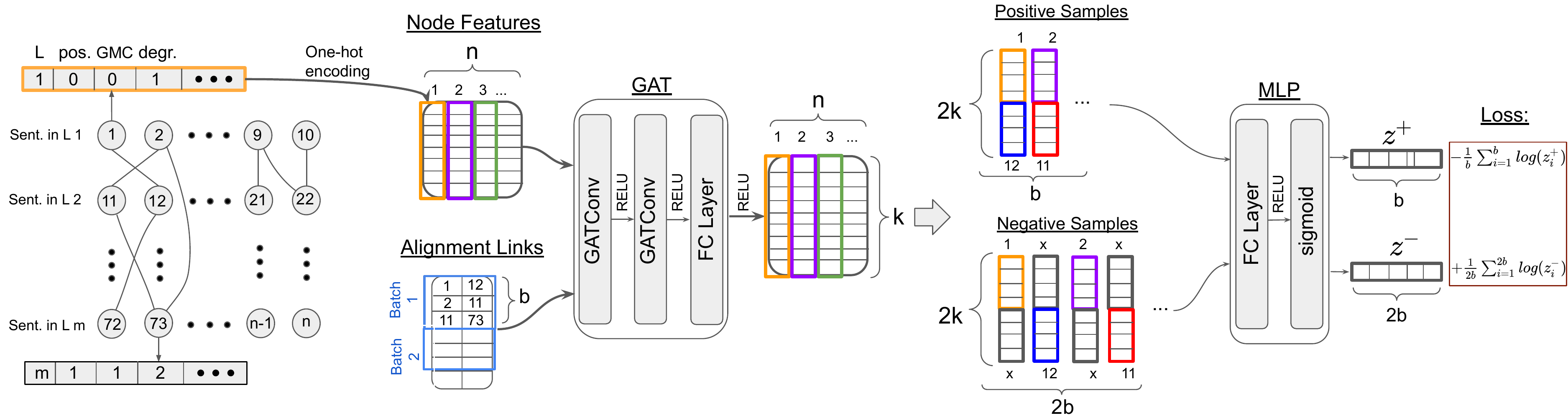}
  \caption{GNN training. At each training step, 
  node features and links of a multiparallel sentence 
  are fed to a graph attention network (GAT) that creates hidden representations 
  for all nodes. On the decoder side, at each step, one batch of 
  alignment links and hidden node representations is used to 
  create positive and negative samples, which are then processed 
  and classified by a multi-layer perceptron (MLP). Parameters of 
  GAT and MLP are updated for each batch. FC = fully connected.
  }
  \label{fig:GNN_model}
\end{figure*}

Our model is inspired by the Graph Auto Encoder (GAE) model of
 \citet{kipf2016variational} for link prediction. 
A GAE model consists of an encoder and a decoder. 
The encoder creates a hidden representation for each node of the graph and 
the decoder predicts the links of the graph given the nodes' representations.
Using the graph of word alignments, the model will learn the word alignment task. 
Therefore it will be able to predict word alignments that are missed by the original 
bilingual word aligner and also detect incorrect alignment edges.

\myblue{We make changes to this model to improve
the model's quality and reduce its computational cost.
We use GATConv layers \citep{velickovic2018graph} for the encoder instead of GCNConv \citep{kipf2017semi} and a more sophisticated
decoder instead of simple dot product for a stronger model.
We also introduce a more efficient training procedure.}

The \textbf{encoder} is a graph attention network (GAT) \citep{velickovic2018graph} with
two GATConv layers followed by a fully connected layer. Layers are
connected by RELU non-linearities.
A GATConv layer computes its output $\mathbf{x}^{\prime}_i$ 
for a node $i$ from its input $\mathbf{x}_{i}$ as
\begin{equation}
\mathbf{x}^{\prime}_i = \alpha_{i,i}\mathbf{W}\mathbf{x}_{i} +
        \sum_{j \in \mathcal{N}(i)} \alpha_{i,j}\mathbf{W}\mathbf{x}_{j},
\end{equation}
where $\mathbf{W}$ is a weight matrix, $\mathcal{N}(i)$ is 
some \textit{neighborhood} of node $i$ in the graph, and $\alpha_{i,j}$
is the attention coefficient indicating the importance of node $j$'s features to node $i$.
$\alpha_{i,j}$ is computed as
\begin{equation}
  \alpha_{i,j} =
  \frac{
  \exp\left(\mathrm{g}\left(\mathbf{a}^{\top}
  [\mathbf{W}\mathbf{x}_i \, \Vert \, \mathbf{W}\mathbf{x}_j]
  \right)\right)}
  {\sum_{k \in \mathcal{N}(i) \cup \{ i \}}
  \exp\left(\mathrm{g}\left(\mathbf{a}^{\top}
  [\mathbf{W}\mathbf{x}_i \, \Vert \, \mathbf{W}\mathbf{x}_k]
  \right)\right)} 
\end{equation}
where $\Vert$ is concatenation,
$g$ is LeakyReLU, and 
$\mathbf{a}$ is a weight vector.
Given the features for the nodes and their alignment edges,
the encoder creates a contextualized hidden representation for each node.

Based on the hidden representations of
two nodes, the \textbf{decoder} predicts whether a link connects them.
The  decoder architecture consists of a fully connected layer,
 a RELU non-linearity
and
a sigmoid layer.

\subsubsection{Training}
\seclabel{sssec:gnn_training}

\myblue{By default, GAE models  are trained using full
batches with 
random negative samples. This approach requires at least
tens of epochs over the training dataset
to converge and a lot  of GPU memory
for graphs as large as ours.
We train our model using mini-batches to decrease memory requirements and improve
the performance. Using our training approach the model converges after one epoch.
We take care to select informative negative samples
(as opposed to random selection)
as described below.}

Figure~\ref{fig:GNN_model} displays our GNN model and the training process.
The training set contains one graph for
each sentence.
Each graph is divided into multiple batches.
Each batch contains a random subset of the graph's edges as positive samples.
The negative samples are created as follows.
Given a sentence $u_1 u_2  \dots{} u_n $
in language $U$ and its translation $v_1 v_2  \dots{} v_m$ in language $V$, 
for each alignment edge $u_i$:$v_j$ in the current batch, 
two negative edges $u_i$:$v_j'$ and $u_i'$:$v_j$
($j'\neq j$, $i'\neq i$) are randomly sampled.

For each training batch, the encoder takes the batch's whole graph (i.e.,
node features for all graph nodes and  all  graph
edges) as input and computes
hidden representations for the nodes.
On the decoder side, for each link between two nodes in the batch,
the hidden representations of the two nodes
are concatenated to create the decoder's input. The decoder's target
is the link class: 1 (resp.\ 0) for positive
(resp.\ negative) links.
We train with a binary classification objective:
         \begin{equation}
  {\cal L} = -\frac{1}{b} \sum_{i=1}^{b}\log(p^+_i) +\frac{1}{2b} \sum_{i=1}^{2b}\log(p^-_i)
\end{equation}
where $b$ is the batch size and $p^+_i$ and $p^-_i$ are the model predictions for 
the $i^{th}$ positive and negative samples within the batch.
Parameters of the encoder and decoder as well as the node-embedding feature layer
are updated after each training step. 

\subsubsection{Node Features}
\seclabel{sssec:nodefeatures}

We use three main types of node features:
(i) graph structural features, (ii) community-based features 
and (iii) word content features.

\textbf{Graph structural features.} 
We use \textit{degree, closeness} \citep{freeman1978centrality}
\textit{, betweenness} \citep{brandes2001faster} \textit{, load} \citep{newman2001scientific}
and \textit{harmonic centrality} \citep{boldi2014axioms}
features as additional information about the graph structure.
These features are continuous numbers, providing information about
 the position and connectivity of the nodes within the graph.
We standardize (i.e., z-score) each feature across all nodes, and
train an embedding of size four for each feature.\footnote{Learning a size-four embedding  instead of a
single number gives the feature 
 a weight similar to
other features -- which have a feature vector of about the
same size.}

\textbf{Community-based features.}
\myblue{One way to incorporate community information into our model is to train the model 
based on a refined set of edges after the community detection step. This approach hobbles the GNN model
by making decisions about many of the edges before the GNN gets to see them. Our 
initial experiments also confirmed that training the GNN
over CD refined edges does not help. Therefore,
we add community information as node features and let the GNN use them to improve its decisions.
}
We use the community detection 
algorithms GMC and LPC (see~\S\secref{ssec:community}) to identify communities in the graph.
Then we represent the community membership information of the nodes as one-hot vectors and
learn an embedding of size~32 for each of the two algorithms.

\textbf{Word content features.} 
We train embeddings for \emph{word position} (size 32)  and  \emph{word
language}  (size 20).
We learn 100-dimensional multilingual 
\emph{word embeddings} using \citet{levy-etal-2017-strong}'s sentence-ID method  on the 
84 PBC languages selected by
\citet{imani2021MPWA}.
Word embeddings serve as initialization and 
are updated during GNN training.

After concatenating these features, each node is represented 
by a 236 dimensional vector that  is then fed to the encoder.

\subsubsection{Inducing Bilingual Alignment Edges}
\seclabel{sssec:inducing_edges}
Given a source sentence $\hat{x} = x_1, x_2, \dots, x_m$ in language 
$X$ and a target sentence $\hat{y} = y_1, y_2, \dots, y_l$ in language $Y$, 
we feed all possible alignment links between source and target to the decoder.
This produces a \myblue{symmetric} alignment probability matrix $S$ 
of size $m\times l$ where
$S_{ij}$ is the predicted alignment probability between words $x_i$ 
and $y_j$. Using these values directly to infer alignment 
edges is usually suboptimal; therefore, more sophisticated 
methods have been suggested
\citep{ayan-dorr-2006-maximum,liang-etal-2006-alignment}.
Here  we propose a new approach:
it combines  \citet{koehn2005GDFA}'s Grow-Diag-Final-And (GDFA) 
with \citet{dou-neubig-2021-word}'s  probability thresholding.
We modify the latter to 
account for the variable size of the probability matrix 
(i.e., length of source/target sentences). 
Our method is not limited to adding new edges to 
some initial bilingual alignments,
a limitation of prior work. 
As we predict each edge independently, some initial links can
be discarded from the final alignment.

We start by creating a set of \textit{forward} (source-to-target) alignment edges and 
a set of \textit{backward} (target-to-source) alignment edges. 
To this end, first, inspired by  probability thresholding  \citep{dou-neubig-2021-word}, we apply softmax to $S$, and zero out probabilities below a 
threshold to get a source-to-target probability matrix $S^{XY}$:
\begin{equation}
S^{XY} = S * (\mbox{softmax}(S) > \frac{\alpha}{l})
\end{equation} 
Analogously, we compute the target-to-source probability matrix $S^{YX}$:
\begin{equation}
S^{YX} = S^\top * (\mbox{softmax}(S^\top) > \frac{\alpha}{m})
\end{equation}
where $\alpha$ is a sensitivity hyperparameter, e.g.,  $\alpha = 1$ means that we pick edges with a probability higher
than average.
We experimentally set $\alpha = 2$. 
Next, from each row of $S^{XY}$ ($S^{YX}$), we pick the cell 
with the highest value (if any exists) and add this edge
to the \textit{forward} (\textit{backward}) set.

We create the final set of alignment edges by applying
the GDFA symmetrization method \cite{koehn2005GDFA} to \textit{forward} 
and \textit{backward} sets. 
The gist of GDFA is to use the intersection of \textit{forward} and \textit{backward} 
as initial alignment edges and add more edges from the union of  
\textit{forward} and \textit{backward} based on a number of heuristics. 
We call this method \textit{TGDFA} (Thresholding GDFA).

We also experiment with combining TGDFA with
the original bilingual GDFA alignments.
We do so by adding bilingual GDFA edges to the union
of  \textit{forward} and \textit{backward} before performing the GDFA heuristics.
We refer to these alignments as \textit{TGDFA+orig.}

\myblue{We evaluate the resulting alignments using $F_1$ score and alignment error rate (AER), the standard metrics in the word alignment literature.}

\begin{table*}[t]
	\small 
	\centering
	\def\tablesep{0.1cm}
	\begin{tabular}{
			@{\hspace{\tablesep}}l@{\hspace{\tablesep}}||
			@{\hspace{\tablesep}}c@{\hspace{\tablesep}}
			@{\hspace{\tablesep}}c@{\hspace{\tablesep}}
			@{\hspace{\tablesep}}c@{\hspace{\tablesep}}
			@{\hspace{\tablesep}}c@{\hspace{\tablesep}}|
			@{\hspace{\tablesep}}c@{\hspace{\tablesep}}
			@{\hspace{\tablesep}}c@{\hspace{\tablesep}}
			@{\hspace{\tablesep}}c@{\hspace{\tablesep}}
			@{\hspace{\tablesep}}c@{\hspace{\tablesep}}|
			@{\hspace{\tablesep}}c@{\hspace{\tablesep}}
			@{\hspace{\tablesep}}c@{\hspace{\tablesep}}
			@{\hspace{\tablesep}}c@{\hspace{\tablesep}}
			@{\hspace{\tablesep}}c@{\hspace{\tablesep}}
			}
		  & \multicolumn{4}{c}{ FIN-HEB } & \multicolumn{4}{c}{ FIN-GRC } & \multicolumn{4}{c}{ ENG-FRA} \\
		  Method & Prec. & Rec. & $F_1$ & AER & Prec. & Rec. & $F_1$ & AER & Prec. & Rec. & $F_1$  & AER \\
		\midrule
		\midrule
		   Eflomal (intersection) & \bfseries0.818 & 0.269          & 0.405           & 0.595          & \bfseries0.897 & 0.506          & 0.647          & 0.353          & \bfseries0.971 & 0.521          & 0.678          & 0.261         \\
		  Eflomal (GDFA)         & 0.508          & 0.448          & 0.476           & 0.524          & 0.733          & 0.671          & 0.701          & 0.300          & 0.856          & 0.710          & 0.776          & 0.221         \\
		\midrule
		\midrule
		 WAdAd (intersection)          & 0.781          & 0.612          & 0.686  & 0.314 & 0.849          & 0.696          & 0.765 & 0.235 & 0.938          & 0.689          & 0.794          & 0.203           \\
		  NMF (intersection)  & 0.780 & 0.576 & 0.663 & 0.337 &    0.864 & 0.669 & 0.754 & 0.248 &    0.948 & 0.624 & 0.753 & 0.245 \\
		  WAdAd (GDFA)            & 0.546          & \bfseries0.693 & 0.611           & 0.389          & 0.707          & \bfseries0.783 & 0.743          & 0.257          & 0.831          & \bfseries0.796 & 0.813 & 0.186 \\
		  NMF (GDFA)                    & 0.548          & 0.646 & 0.593           & 0.407          & 0.720          & 0.759 & 0.739          & 0.261          & 0.844          & 0.767 & 0.804 & 0.195  \\
    \midrule
    GNN (TGDFA)   & 0.811          & 0.648 & \bfseries0.720           & \bfseries0.280         & 0.845          & 0.724 & \bfseries0.780          & \bfseries0.220          & 0.926         & 0.711 & 0.804 & 0.192 \\
    GNN (TGDFA+orig) & 0.622         &  0.683 & 0.651          & 0.349          & 0.738          & 0.780 & 0.758          & 0.242          & 0.863         & 0.789 & \bfseries0.824 & \bfseries0.174  \\
      \midrule\midrule
	\end{tabular}
	\caption{Word alignment results on PBC for GNN
	and baselines.
	The best result in each column is in bold. GNN
  outperforms the baselines as well as the  graph algorithms WAdAd and NMF
  on $F_1$ and AER.}
	\tablabel{res_pbc}
\end{table*}

\subsection{Annotation Projection}

Annotation projection automatically creates
linguistically annotated corpora for low-resource languages.
A model trained on data with
``annotation-projected'' labels can perform better than a
completely unsupervised method.
Here, we focus on universal part-of-speech (UPOS) tagging 
\citep{petrov-etal-2012-universal} 
for the low resource 
target language Yoruba; this language 
only has a small set of annotated sentences in Universal
Dependencies \citep{nivre-etal-2020-universal} and has poor
POS results in unsupervised settings \citep{kondratyuk-straka-2019-75}.

The quality of the target annotated corpus
depends on the quality of the annotations in the source languages and the
quality of the word alignments between sources and target.
We use  the Flair \citep{akbik-etal-2019-flair} POS taggers
for three high resource languages, English, German and French \citep{akbik-etal-2018-contextual},
to annotate 30K verses whose
Yoruba translations 
are available in PBC. We then transfer the POS tags from source to target
using three different approaches:
(i) We directly transfer annotations from English to the target.
(ii) For each word in the target, we get its Eflomal bilingual alignments 
in the three source languages and predict the majority POS
to annotate the target word. 
(iii)
The same as in (ii), but we use 
our GNN (TGDFA) alignments (instead of Eflomal alignments) to project from source to target.
In all three approaches, we discard any target sentence from the POS tagger training data if more than 50\% of its words are annotated with the "X" (other) tag.

We train a Flair SequenceTagger model
on the target annotated data 
using mBERT embeddings \cite{devlin-etal-2019-bert} and 
evaluate on Yoruba test from Universal 
Dependencies.\footnote{\url{https://universaldependencies.org/}}

\section{Experimental Setup}

\subsection{Word Alignment Datasets}
\seclabel{ssec:align_datasets}

Following \citet{imani2021MPWA}, we use PBC, a
multiparallel corpus of
1758 sentence-aligned editions of the Bible in 1334 languages.

\textbf{\myblue{Evaluation data.}}
For our main evaluation, we use the two word alignment gold
datasets
for PBC published by 
\citet{imani2021MPWA}: Blinker \citep{melamed1998blinker} 
and HELFI \citep{yli-jyra-etal-2020-helfi}.
\textbf{The HELFI dataset} contains the Hebrew Bible, Greek New Testament 
and their translations into Finnish. 
For HELFI,  we use \citet{imani2021MPWA}'s train/dev/test splits.
\textbf{The Blinker dataset} provides word level alignments between English and French 
for 250 Bible verses. 

\textbf{\myblue{Training data.}}
The graph algorithms used by \citet{imani2021MPWA} 
operate on each multiparallel sentence separately. In contrast, our approach allows for an inductive 
setting where a model is trained on a training set, accumulating knowledge from multiple multiparallel sentences.
We combine the verses in the training sets of Finnish-Hebrew and
Finnish-Greek for a combined training set size of 
24,159.\footnote{Note that we do not use any gold alignments for training the GNN. Using the verses from HELFI train split as our training set is for convenience. Our ablation experiment (Figure \ref{fig:training_size}) show that a smaller subset of the training set is sufficient to achieve good performance}

\subsection{Initial Word Alignments}
\seclabel{ssec:init_aligns}

We use the \eflomal{} statistical word aligner to obtain bilingual alignments. 
We train it for every language pair in our experiments.
We do not consider SimAlign \citep{jalili-sabet-etal-2020-simalign} since it is 
shown to perform poorly for languages whose representations in the 
multilingual pretrained language model are of low quality.
We use \eflomal{} asymmetrical alignments post-processed with the intersection heuristic to get  high precision bilingual 
alignments as input to the GNN. 
We use the same subset of 84 languages
as \citet{imani2021MPWA}.

\subsection{Training Details}

We use 
PyTorch Geometric\footnote{\url{pytorch-geometric.readthedocs.io}}
to construct and train the GNN.
The model's hidden layer size is 512 for both GATConv and Linear layers.
We train for one epoch on the training set --
a small portion of the training set is enough
to learn good embeddings
(see  \secref{train_size}).
For training, we use a batch size of~400 and learning rate of .001 with 
AdamW \cite{loshchilov2017decoupled}. 
\myblue{The whole training process takes less
than 4 hours on a GeForce GTX 1080 Ti and the inference time is on the
order of milliseconds per sentence.}

\section{Experiments and Results}

\subsection{Multiparallel corpus results}

Table \ref{tab:res_pbc} shows results on Blinker and HELFI for our GNNs and the baselines:
 bilingual alignments and two graph-based algorithms WAdAd and NMF from \citet{imani2021MPWA}.
Our GNNs  yield a better trade-off between precision 
and recall, most likely thanks to their ability to remove edges, and
achieve the best $F_1$ and AER on all three datasets,
outperforming WAdAd and NMF.

GNN (TGDFA) achieves the best results on HELFI  (FIN-HEB, FIN-GRC)
while GNN (TGDFA+orig) is best on Blinker (ENG-FRA).
As argued in \citet{imani2021MPWA}, this is mostly due to the different 
ways these two datasets were annotated. 
Most HELFI alignments are one-to-one, while many Blinker  alignments 
are many-to-many:  phrase-level alignments where every word in a source
phrase is aligned with every word in a target phrase.
This suggests that one can choose between GNN (TGDFA) and GNN (TGDFA+orig) 
based on the desired characteristics of the alignment. 

\subsubsection{Effect of Training Set Size}
\begin{figure}
  \centering
  \includegraphics[width=\columnwidth]{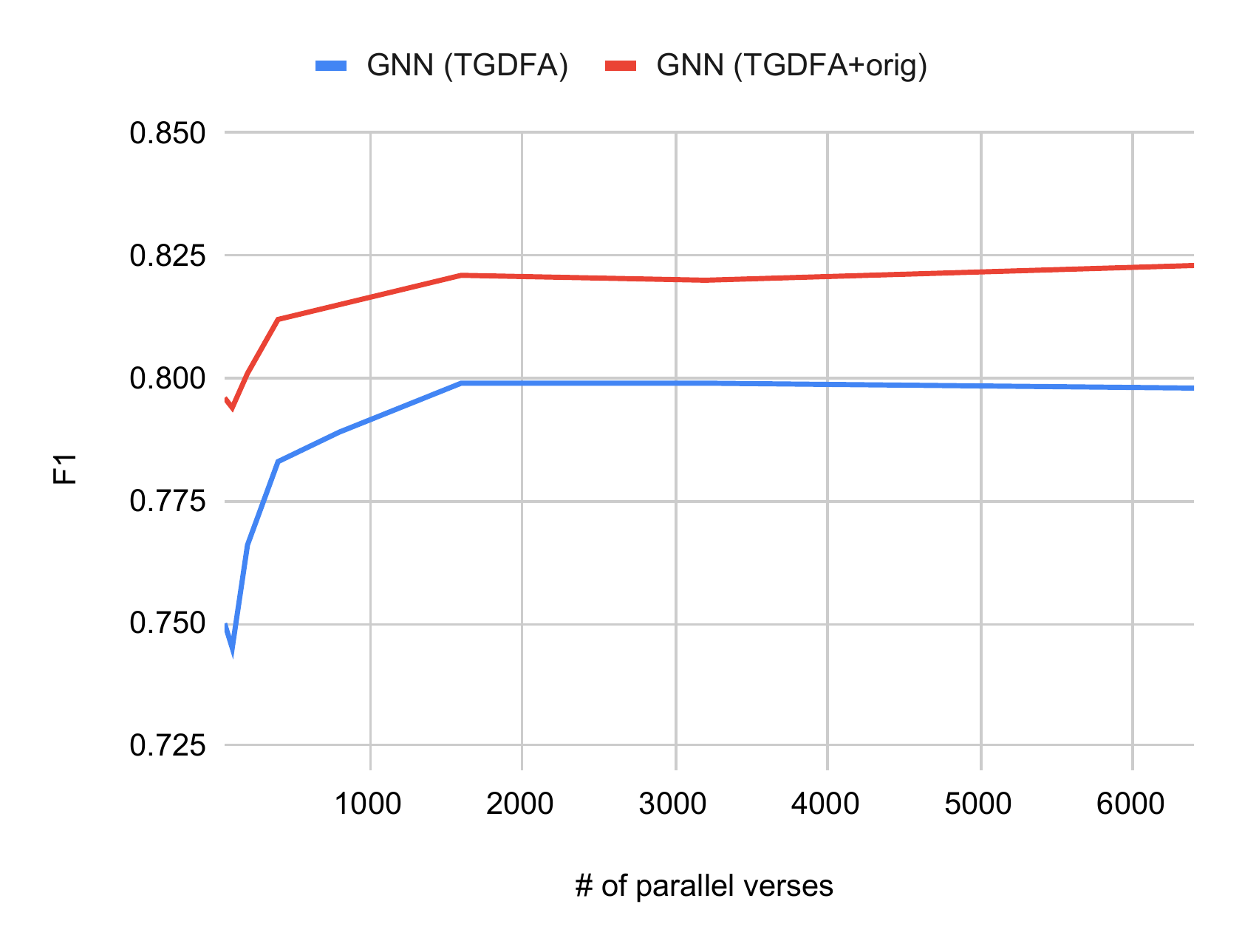}
  \caption{$F_1$ of GNN (TGDFA) and GNN (TGDFA+orig) on  Blinker 
  as a function of train size
  }
  \label{fig:training_size}
\end{figure}

\seclabel{train_size}
To investigate the effect of training set size,
we train the GNN on subsets of our training data with increasing sizes.
Figure~\ref{fig:training_size} shows results.
Performance improves fast until around 2,000 verses; 
then it stays mostly constant. 
Using more than 6,400 samples 
does not change the performance at all. 
Therefore, in the other experiments we use 6,400 randomly sampled verses 
from the training set to train  GNNs.

\subsubsection{Ablation Experiments}

To examine the importance of  node features,
we ablate
{language}, {position}, 
{centrality}, {community} and 
{word embedding}  features.
Table~\ref{tab:ablation} shows that
removal of  graph structural features drastically 
reduces  performance. Community features and 
language information are also important.
Removal of word position information and 
word embeddings -- which store semantic information about
words -- has the least effect. Based on these results, it can 
be argued that
the lexical information 
contained in the initial alignments and in the community 
features provides a strong signal regarding word relatedness. 
The novel information that is crucial is about the overall 
graph structure which goes beyond the local word
associations that are
captured by word position and word embeddings.

\begin{figure}[ht]
  \centering
  \includegraphics[width=\columnwidth]{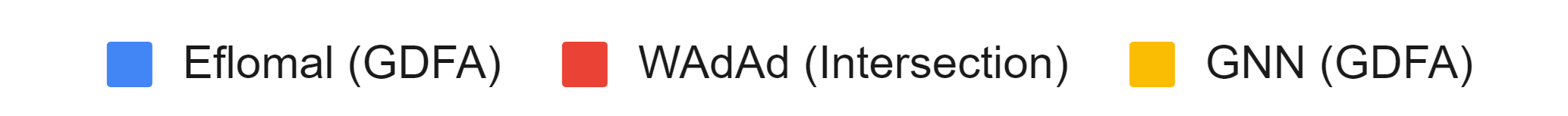}
  \\
  \centering
  \subfloat[ENG-FRA]{\label{fig:mdleft}{\includegraphics[width=0.5\columnwidth]{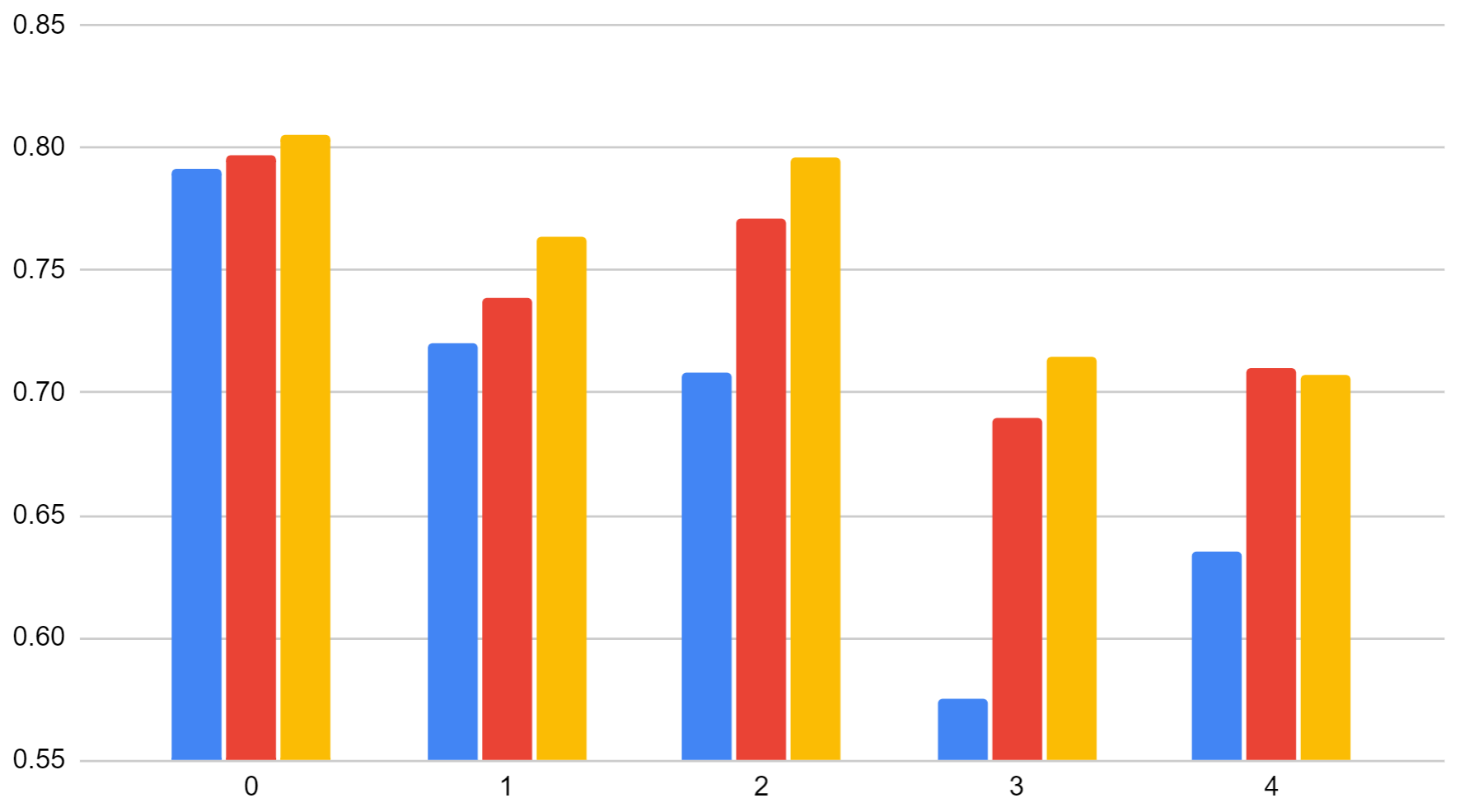}}}\hfill
  \subfloat[FIN-HEB]{\label{fig:mdright}{\includegraphics[width=0.5\columnwidth]{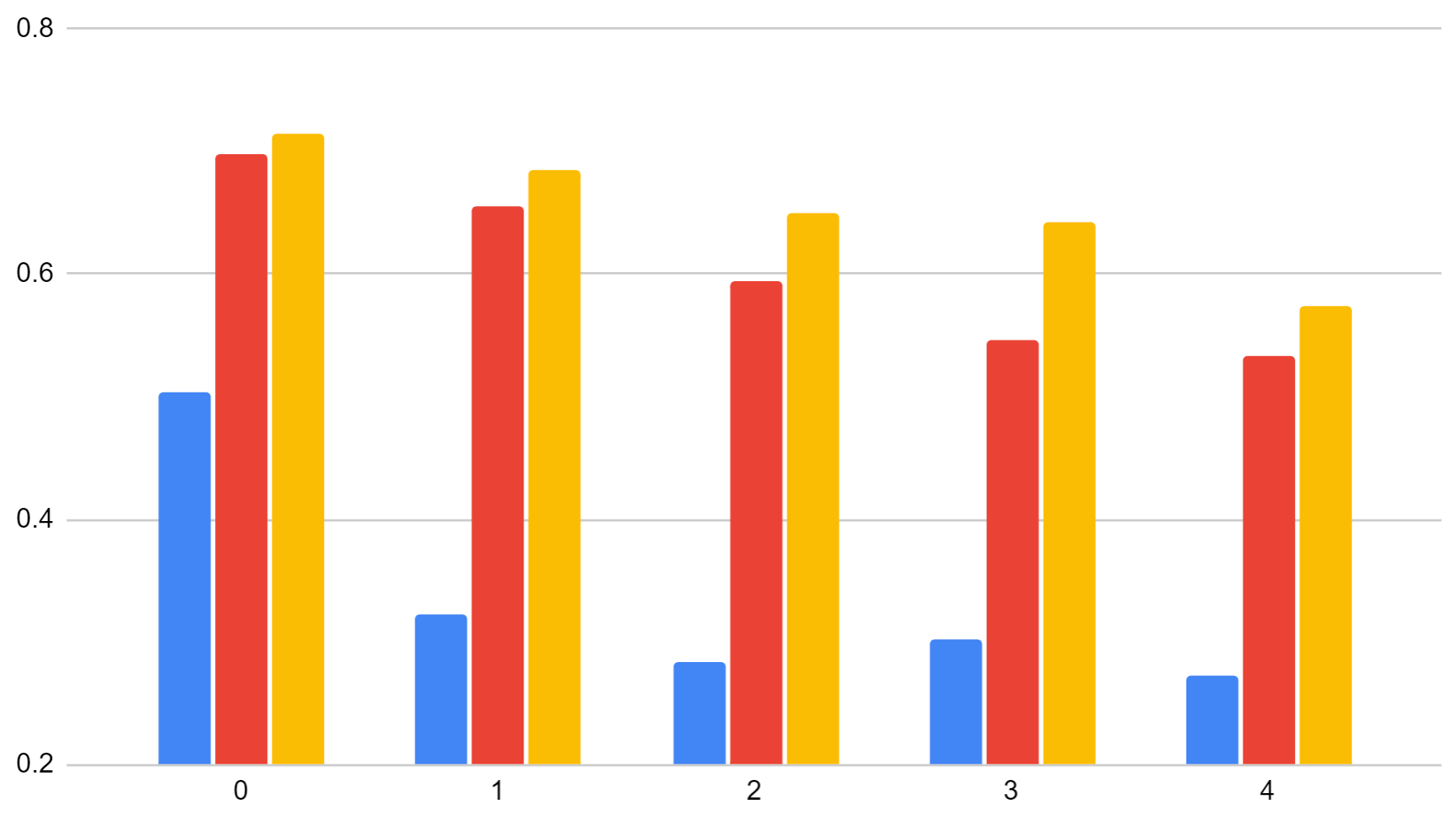}}}
  \caption{$F_1$  for different frequency bins.}
  \label{fig:diff}
\end{figure}

\myblue{
\subsubsection{Effect of Word Frequency}
We investigate the
effect of word frequency on
alignment performance where
frequency is calculated based on the source word in the PBC;  the first bin has the highest frequency.
Figure~\ref{fig:diff} shows that the performance of Eflomal
drops with frequency
and it struggles to align very rare words.
In contrast, GNN  is not affected by  word frequency as severely  
and its performance gains are even greater for rare words. WAdad which is the multilingual baseline
from \cite{imani2021MPWA} has the same trend as the GNN
method, but the GNN method is more robust.
}

\begin{table}[t]
	\small 
	\centering
	\def\tablesep{0.1cm}
	\begin{tabular}{
			@{\hspace{\tablesep}}l@{\hspace{\tablesep}}||
			@{\hspace{\tablesep}}r@{\hspace{\tablesep}}
			@{\hspace{\tablesep}}r@{\hspace{\tablesep}}
			@{\hspace{\tablesep}}r@{\hspace{\tablesep}}
			}
		 & FIN-HEB  & FIN-GRC & ENG-FRA \\
		\midrule
		\midrule
    GNN (TGDFA)   &  0.720    & 0.780    & 0.804 \\ 
    \midrule
$\neg$ language    &	-0.323   &	-0.280  &	-0.370 \\
$\neg$ position    &	-0.068  &	-0.045 &	-0.066  \\
$\neg$ centrality   & -0.636  & -0.730  &	-0.772  \\
$\neg$ community    & -0.204  & -0.238  & -0.253  \\
$\neg$ word-embedding   & -0.139  & -0.103 & -0.129 \\
    \midrule
    \midrule
    GNN (TGDFA+orig) &  0.651     & 0.758    & 0.824  \\		
    \midrule
$\neg$ language  &	-0.238  &	-0.077  &	-0.162	\\		
$\neg$ position  &	-0.088  &	+0.029	& -0.032	 \\	
$\neg$ centrality  & -0.556  & -0.530  & -0.617  \\	
$\neg$ community  & -0.156 & -0.039 & -0.083 \\	
$\neg$ word-embedding  & -0.135  & +0.002 & -0.058  \\		
    \midrule
    \midrule
	\end{tabular}
	\caption{$F_1$ for GNNs
	and $\Delta F_1$ for five ablations}
	\tablabel{ablation}
\end{table}

\subsection{Annotation Projection}

Table~\ref{tab:annotation_projection} presents 
accuracies for POS tagging in Yoruba.
Unsupervised baseline performance is 50.86\%.
Supervised training using pseudo-labels mostly outperforms 
the unsupervised baseline.
Projecting the majority POS labels to Yoruba improves over
projecting English labels.
Using the GNN model to project labels works best and
outperforms 
\eflomal-GDFA-majority (resp.\ the unsupervised baseline) by 5\%
(resp.\ 15\%) 
absolute improvement.

\begin{table}[t]
	\small 
	\centering
	\def\tablesep{0.1cm}
	\begin{tabular}{
			@{\hspace{\tablesep}}l@{\hspace{\tablesep}}||
			@{\hspace{\tablesep}}c@{\hspace{\tablesep}}
			}
		Model & Yoruba YTB \\
		\midrule
		\midrule
    Unsupervised \cite{kondratyuk-straka-2019-75}   &  50.86   \\ 
    \midrule
    \eflomal{} Inter -- eng   &  43.45  \\ 
    \eflomal{} GDFA -- eng  &  55.13  \\ 
    \midrule
    \eflomal{} Inter -- majority  &  54.13  \\ 
    \eflomal{} GDFA -- majority  &  60.27  \\ 
    \midrule
    GNN (TGDFA) -- majority  & \bfseries 65.74  \\ 
    GNN (TGDFA+orig) -- majority  &  64.55  \\
    \midrule
    \midrule
	\end{tabular}
	\caption{POS tagging with annotation projection for
	Yoruba.
        Apart from ``Unsupervised'',
	all lines
	show a sequence tagger trained on pseudo-labels
	induced by word alignments.
        GNN-based pseudo-labels outperform prior work by
	5\% absolute.}
	\tablabel{annotation_projection}
\end{table}

\section{Related Work}

\textbf{Bilingual Word Aligners.} 
Much work on bilingual word alignment is based on probabilistic models, mostly implementing variants of the
IBM models of \citet{brown-etal-1993-mathematics}: e.g.,
Giza++ \citep{och-ney-2003-systematic}, fast-align \citep{dyer-etal-2013-simple} and 
\eflomal{} \citep{ostling2016efficient}. 
More recent work, including SimAlign \citep{jalili-sabet-etal-2020-simalign} 
and SHIFT-ATT/SHIFT-AET \citep{chen-etal-2020-accurate}, 
uses pretrained neural language and machine translation 
models.
Although neural models achieve superior performance compared to 
statistical aligners,
they can only be used for
fewer than two hundred 
high-resource languages that are supported by multilingual language models like BERT 
\citep{devlin-etal-2019-bert} and 
XLM-R \citep{conneau-etal-2020-unsupervised}. This makes statistical models 
the only option for the majority of the world's languages.

\textbf{Multiparallel Corpora.}
Prior applications of  using multiparallel corpora
include
reliable translations from small datasets \citep{cohn-lapata-2007-machine},
and phrase-based machine translation (PBMT) \citep{kumar-etal-2007-improving}.
Multiparallel corpora are also used for language comparison
 \citep{mayer-cysouw-2012-language}, typological
studies \citep{ostling-2015-word, asgari-schutze-2017-past} 
and PBMT \citep{nakov2012improving, bertoldi2008phrase, dyer-etal-2013-simple}.
\citet{imanigooghari-etal-2021-parcoure} provide a tool to browse a word-aligned 
multiparallel corpus, which can be used for
the comparative study of languages and for error analysis in
machine translation.

To the best of our knowledge \citet{lardilleux2008truly} and \citet{ostling-2014-bayesian}\footnote{\url{github.com/robertostling/eflomal}}
 are the only word alignment methods designed for multiparallel corpora.
However, the latter method is outperformed by \eflomal{}  \cite{ostling2016efficient},
a bilingual method from the same author. 
Recently, \citet{imani2021MPWA} proposed MPWA,
which we use as our baseline.

\textbf{Graph Neural Networks (GNNs)}
have been used to address many problems
that are inherently  graph-like such as traffic networks, 
social networks, and physical and biological systems \cite{liu2020introduction}.
GNNs achieve impressive performance in many domains, including
social networks \citep{wu2020graphSocial} and
natural science \citep{sanchez18GNNPhysics} 
as well as NLP tasks like
sentence classification \citep{huang-etal-2020-syntax},
question generation \citep{pan-etal-2020-semantic},
summarization \citep{fernandes2019summarization}
and
derivational morphology
\citep{hofmannsp20}.

\section{Conclusion and Future Work}

We introduced
graph neural networks
and community detection algorithms 
for multiparallel word alignment.
By
incorporating
signals from diverse sources as node features, including
community features,
our GNN 
model outperformed the baselines and prior work, 
establishing new state-of-the-art results on  three 
PBC gold standard datasets.
We also showed that our GNN model improves downstream 
task performance in low-resource languages through annotation 
projection.

 We have only used node features to provide signals 
 to  GNNs. In the future, other signals can be added in the form 
 of edge features to further boost the performance.
 
\section{Acknowledgments} 

This work was supported by the European Research Council
(ERC, Grant No.\ 740516) and the German Federal Ministry of Education and
Research (BMBF, Grant No.\ 01IS18036A). We
also thank the anonymous reviewers for their constructive
comments.

\bibliography{anthology,custom}
\bibliographystyle{acl_natbib}

\appendix

\section{Appendix}
\label{sec:appendix}

\subsection{Languages}

\begin{table*}[t]
	\centering
	\def\tablesep{0.1cm}
	\begin{tabular}{llllllll}
		
Afrikaans & Albanian & Arabic & Armenian & Azerbaijani & Bashkir \\ 
Basque & Belarusian & Bengali & Breton & Bulgarian & Burmese \\ 
Catalan & Cebuano & Chechen & Chinese & Chuvash & Croatian \\ 
Czech & Danish & Dutch & English & Estonian & Finnish \\
French & Georgian & German & Greek & Gujarati & Haitian \\ 
Hebrew & Hindi & Hungarian & Icelandic & Indonesian & Irish \\ 
Italian & Japanese & Javanese & Kannada & Kazakh & Kirghiz \\ 
Korean & Latin & Latvian & Lithuanian & Low Saxon & Macedonian \\ 
Malagasy & Malay & Malayalam & Marathi & Minangkabau & Nepali \\ 
Norwegian (B.) & Norwegian (N.) & Punjabi & Persian & Polish & Portuguese \\ 
Punjabi & Romanian & Russian & Serbian & Slovak & Slovenian \\ 
Spanish & Swahili & Sundanese & Swedish & Tagalog & Tajik \\ 
Tamil & Tatar & Telugu & Turkish & Ukrainian & Urdu \\ 
Uzbek & Vietnamese & Waray-Waray & Welsh & West Frisian & Yoruba \\
	
	\end{tabular}
	\caption{List of the 84 languages we used in our experiments.}
	\tablabel{langauges}
\end{table*}


\end{document}